\PassOptionsToPackage{numbers}{natbib}
\documentclass[12pt]{article}

\usepackage[preprint]{neurips_2025}  
\usepackage{array}
\usepackage{booktabs}
\usepackage{amssymb}
\usepackage{pifont}

\usepackage{silence}
\usepackage[protrusion=false, expansion=false]{microtype}

\renewcommand{\textsc}[1]{{\fontsize{0.85em}{1em}\selectfont\uppercase{#1}}}

\usepackage{hyperref}
\usepackage{url}
\usepackage{booktabs}
\usepackage{amsfonts}
\usepackage{nicefrac}
\usepackage{xcolor}
\usepackage{parskip}
\usepackage{setspace}
\usepackage{titlesec}

\titlespacing*{\section}{0pt}{0.3em plus 0.1em minus 0.1em}{0.3em}
\titlespacing*{\subsection}{0pt}{0.3em plus 0.1em minus 0.1em}{0.3em}
\titlespacing*{\subsubsection}{0pt}{0.8em}{0.5em}

\titleformat{\section}
  {\normalfont\fontsize{14pt}{17pt}\bfseries}
  {\thesection}{1em}{}

\titleformat{\subsection}
  {\normalfont\fontsize{12pt}{15pt}\bfseries}
  {\thesubsection}{1em}{}

\titleformat{\subsubsection}
  {\normalfont\fontsize{11pt}{14pt}\bfseries}
  {\thesubsubsection}{1em}{}

\title{Time-Scaling Is What Agents Need Now}
\usepackage{graphicx}

\author{
  \textbf{Zhi Liu} \quad \textbf{Guangzhi Wang} \\[0.5em]
  CareerInternational Research Team \\[0.5em]
}

\begin{document}
\maketitle

\begin{abstract}
Early artificial intelligence paradigms exhibited separated cognitive functions: Neural Networks focused on "perception-representation," Reinforcement Learning on "decision-making-behavior" \cite{rittlejohnson1998strategic}, and Symbolic AI on "knowledge-reasoning." With Transformer-based large models and world models \cite{norman1993things,anderson1990adaptive}, these paradigms are converging into cognitive agents with closed-loop "perception-decision-action" capabilities.

Humans solve complex problems under limited cognitive resources through temporalized sequential reasoning \cite{simon1978information}. Language relies on problem space search for deep semantic reasoning \cite{newell1972human}. While early large language models (LLMs) could generate fluent text, they lacked robust semantic reasoning capabilities. Prompting techniques like Chain-of-Thought (CoT) \cite{wei2022chain} and Tree-of-Thought (ToT) \cite{yao2023tree} extended reasoning paths by making intermediate steps explicit. Recent models like DeepSeek-R1 enhanced performance through explicit reasoning trajectories \cite{deepseekr1}. However, these methods have limitations in search completeness and efficiency.

This highlights the need for "Time-Scaling"--the systematic extension and optimization of an agent's ability to unfold reasoning over time. Time-Scaling refers to architectural design utilizing extended temporal pathways, enabling deeper problem space exploration, dynamic strategy adjustment, and enhanced metacognitive control, paralleling human sequential reasoning under cognitive constraints. It represents a critical frontier for enhancing deep reasoning and problem-solving without proportional increases in static model parameters. Advancing intelligent agent capabilities requires placing Time-Scaling principles at the forefront, positioning explicit temporal reasoning management as foundational.

\end{abstract}

\section{Background and Related Work}

\subsection{Cognitive Psychology}

Cognitive psychology, based on the information processing model, emphasizes that human thinking processes can be analogized to computer information processing mechanisms \cite{anderson1982acquisition}. This mechanism typically includes three interconnected functional modules—perception, cognition, and action—that work together to receive, process, and respond to external information.

\subsubsection{Problem Space Theory}

\begin{figure}[htbp] 
	\centering
	\includegraphics[width=0.9\linewidth]{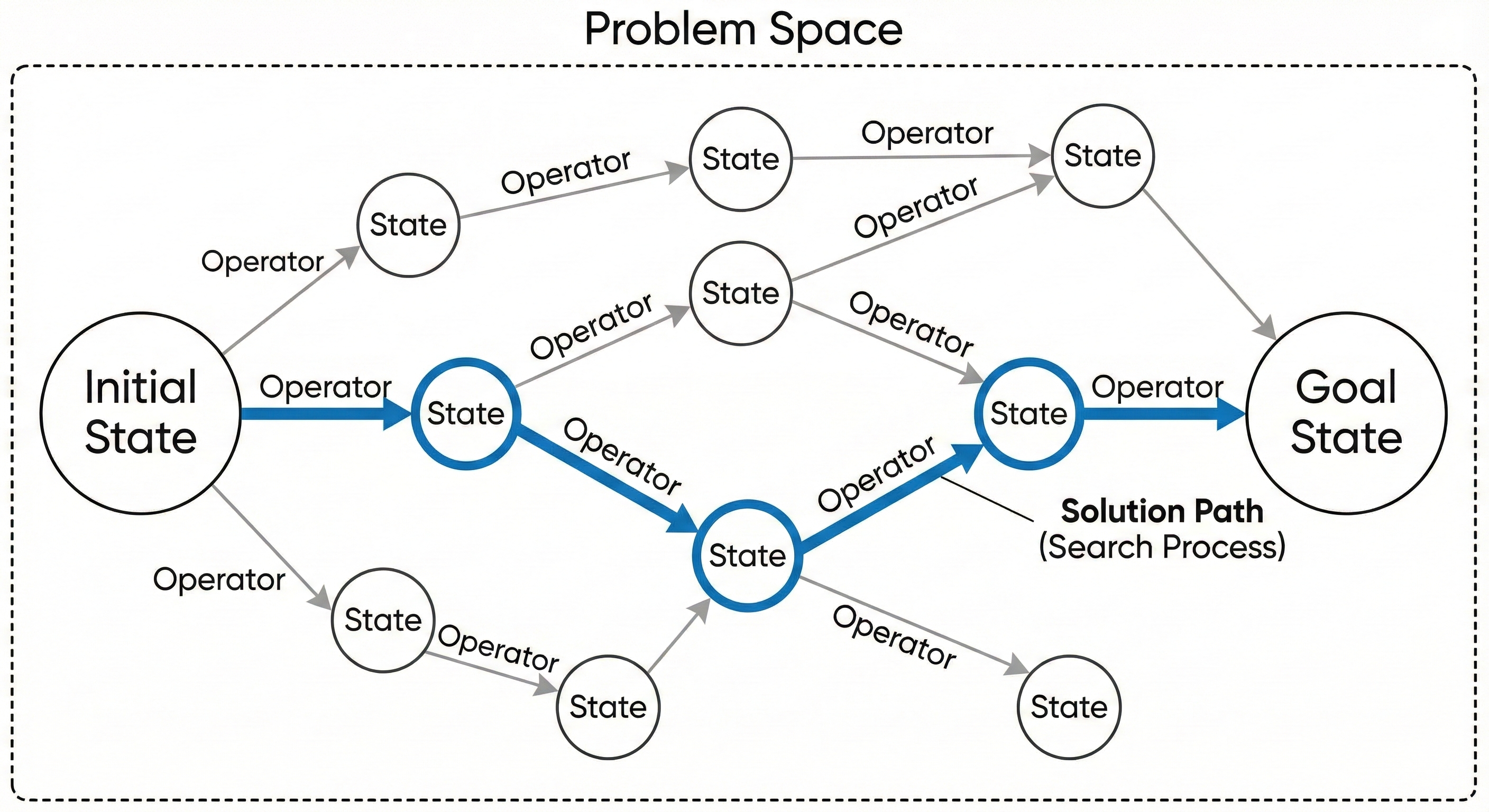}
	\caption{Problem Space Theory}
	\label{fig:problem-space}
\end{figure}

Newell and Simon \cite{newell1972human} proposed the Problem Space Theory, which formalizes problem solving as a triplet: initial state, operators, and goal state. Problem solving is the process of searching for a path from the initial state to the goal state through a series of operations in the state space (Figure~\ref{fig:problem-space}).

The way a problem is stated determines the representation scope and operator availability of the problem space \cite{simon1978information}. Different formulations guide individuals to construct different mental models, thereby affecting reasoning paths and strategy selection. For example, expressing a math problem as "find the sum" versus "find the pattern" will activate completely different problem-solving strategies.

Building on this view, Problem Space Theory further emphasizes how bounded rationality \cite{simon1957models} constrains search: limited cognitive resources typically allow only a partial unfolding of the problem space, thereby prompting the reliance on heuristics \cite{gigerenzer2011heuristic}. In its most general form, a "problem" is thus a unified representational construct for diverse cognitive activities—including planning, reasoning, learning, control, and understanding. This generality establishes Problem Space Theory as a cross-disciplinary foundation, informing models of human thought in psychology and underpinning key AI paradigms such as state-space search, reinforcement learning, and automated planning. For example, the core challenge in reinforcement learning—finding a policy that maximizes cumulative reward—can be directly formulated as an optimal path search from an initial to a goal state within a problem space.

\subsubsection{Problem-Solving Strategies}

Problem-solving strategies are typically divided into two levels: basic strategies and advanced strategies \cite{mayer1992thinking}.

\begin{itemize}
\item \textbf{Basic problem-solving strategies} typically focus on \emph{well-structured problem spaces}, where initial state, goal state, and operators are clearly defined, and the solution space is enumerable (such as classic puzzles, mathematical proofs, etc.) \cite{newell1972human}. Typical methods include means-ends analysis, generate-and-test, backtracking, working backward, and hill climbing. These strategies can often be formalized as symbolic computational models, such as GPS (General Problem Solver) and Soar, demonstrating high efficiency on rule-clear tasks.

\item \textbf{Advanced strategies} address \emph{open or ambiguous tasks}, emphasizing meta-control and reflection of the problem-solving process. Representative forms include analogical reasoning \cite{gentner1983structure}, strategy switching, and insight (Aha Moment) \cite{ohlsson1992information}. These methods embody the metacognitive mechanism of "thinking about one's own thinking" \cite{flavell1979metacognition}.
\end{itemize}

\begin{figure}[htbp]
	\centering
	\includegraphics[width=0.7\linewidth]{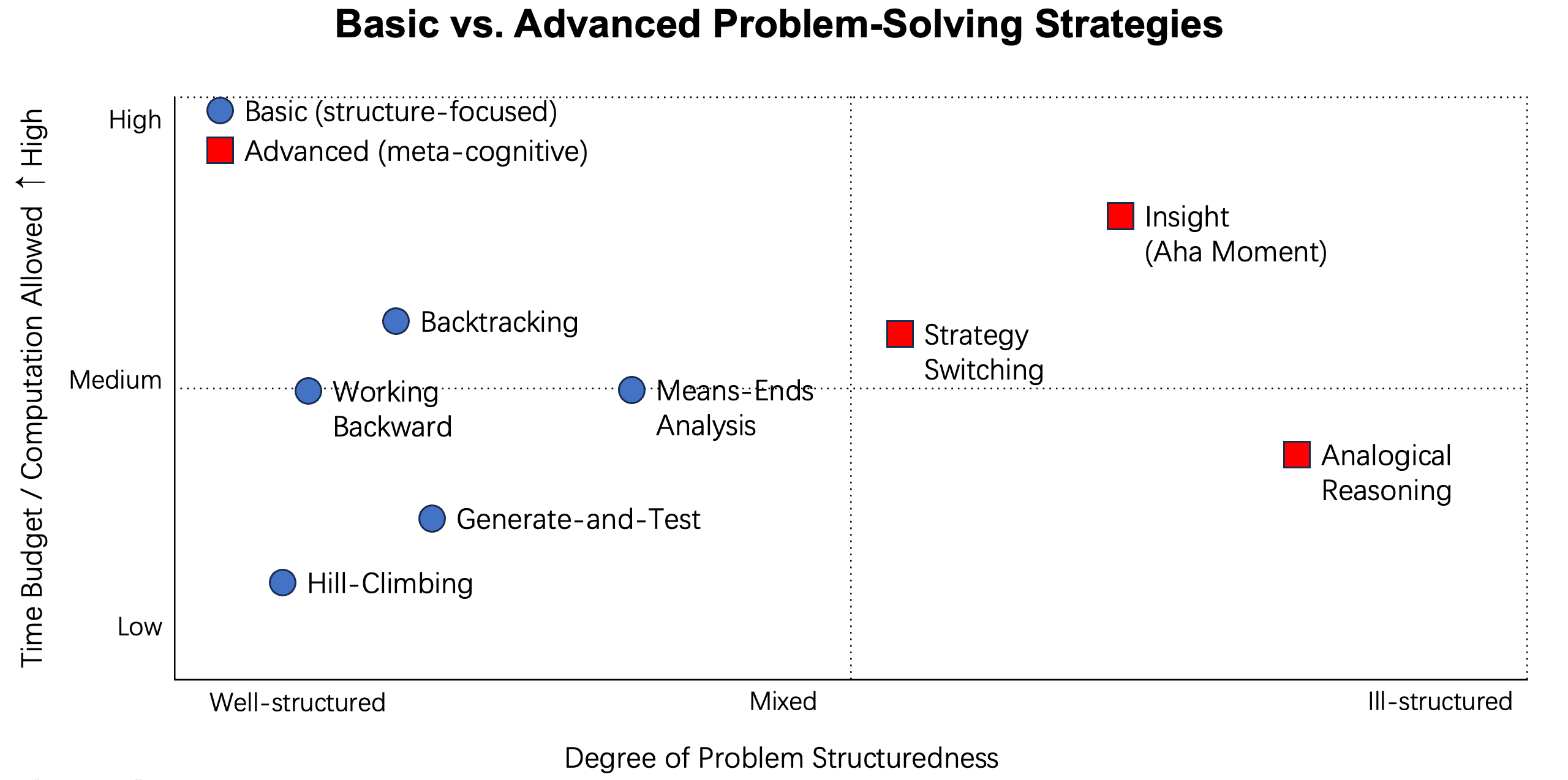}
	\caption{Distribution of basic (blue circles) and advanced (red squares) problem-solving strategies on the two-dimensional coordinate of "problem structuredness" and "time/computation budget." The horizontal axis indicates less structured problems toward the right; the vertical axis indicates more available reasoning steps or computational power upward.}
	\label{fig:strategy-distribution}
\end{figure}

Different problems require different problem-solving strategies. We project common strategies onto a plane composed of two key dimensions—"problem structuredness" and "time budget" (Figure~\ref{fig:strategy-distribution})—to reveal which strategies are more suitable for which situations:

This coordinate-based perspective reveals a natural \emph{temporal scaling pathway}: in resource-constrained scenarios, systems tend to first adopt low-cost, heuristic-driven basic strategies; as allowed reasoning steps or compute increase, they can gradually switch to more complete or creative advanced strategies without an immediate expansion of the model's parameters.

Advanced strategies are deemed "advanced" precisely because they involve explicit awareness of the current cognitive state, dynamic reorganization of solution paths, and strategic utilization of past failures. In contrast to the local, linear, and mechanistic operations of basic strategies, advanced strategies exhibit greater discontinuity, holism, and a nonlinear structure, enabling strong adaptability in task environments characterized by structural incompleteness, ambiguous goals, and multiple potential solution paths.

Notably, recent advances in reasoning-augmented large language models have begun to exhibit explicit emulation of certain advanced problem-solving strategies. For instance, DeepSeek-R1 demonstrates adaptive strategy adjustment, reflective restructuring, and behaviors approximating ``Aha moments'' during its reasoning process. Such mechanisms, through multi-step reasoning and the explicit externalization of intermediate cognitive states, indicate that models are developing rudimentary capabilities for strategic self-regulation when confronting complex tasks. This can be viewed as an early manifestation of advanced problem-solving strategies in artificial intelligence systems, holding significant implications for enhancing large model performance on open-ended tasks and along the trajectory toward artificial general intelligence.

As Newell et al. (1959) observed: ``A genuine problem-solving process involves the repeated use of available information to initiate exploration. This process progressively reveals additional information until a method for finding the solution is ultimately discovered.'' This perspective foreshadows that the transition from basic to advanced strategies essentially represents a cognitive evolutionary process---from search to heuristics, from mechanical execution to strategic introspection.

\subsubsection{Strategy Combination and Scheduling}

A single problem-solving strategy typically only covers specific types of problem structures and temporal constraints. Consequently, in real-world complex tasks, both humans and intelligent agents tend to dynamically combine multiple strategies to adapt to the non-stationarity of the problem space and variations in the available time budget \cite{polya1945how, card1983psychology, rittle1999strategic}. Such strategic combination reflects both the hierarchical nature of problem-solving and the capacity of metacognition to orchestrate the overall process.

\begin{figure}[htbp]
	\centering
	\includegraphics[width=0.3\linewidth]{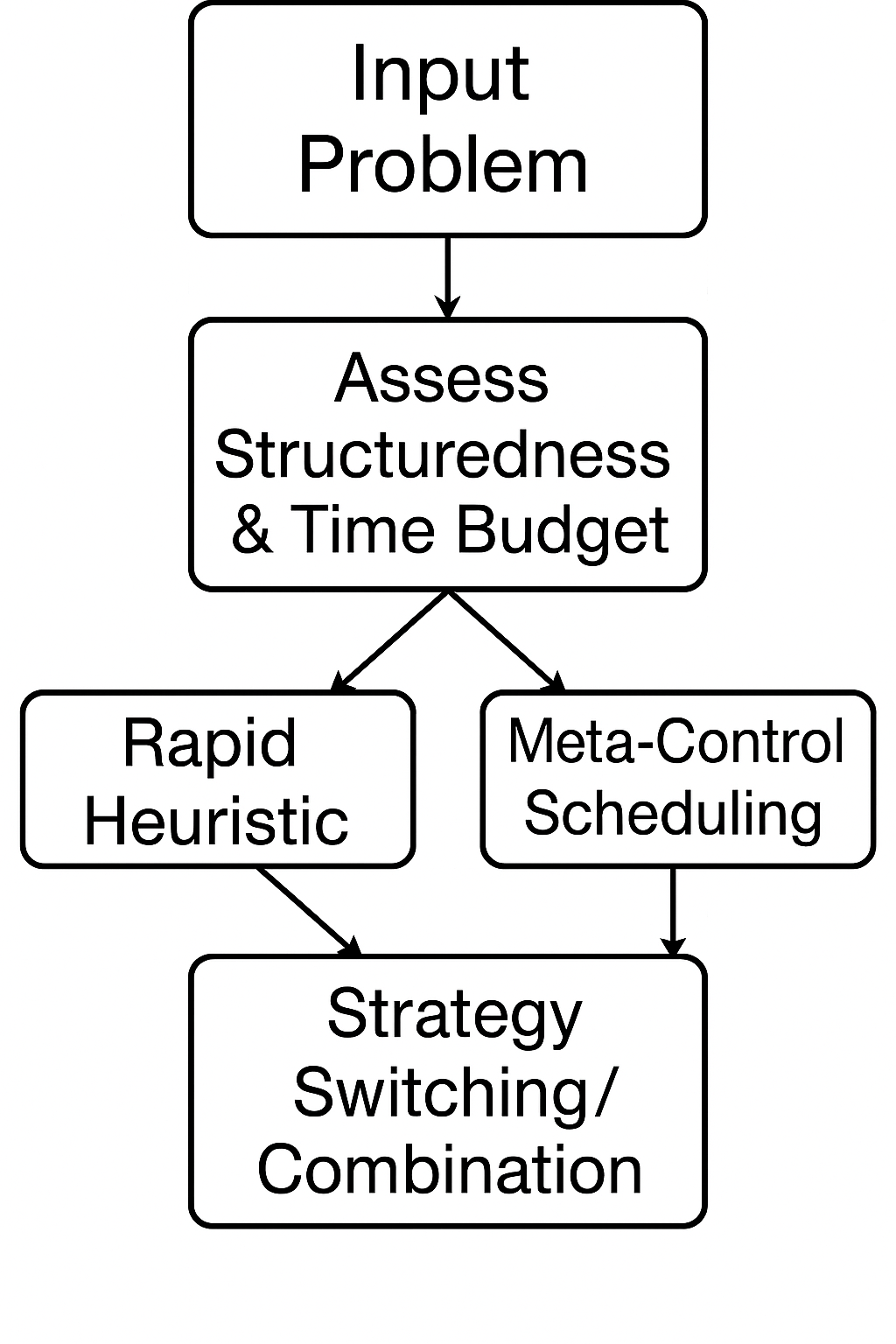}
	\caption{Flowchart of strategy combination}
	\label{fig:strategy-flowchart}
\end{figure}

\begin{itemize}
\item \textbf{Sequential combination}: Strategies are attempted in order from low to high heuristic strength and computational cost. For example, the system first applies hill climbing; if it becomes trapped in a local optimum, it switches to backtracking or breadth-first search. This combination paradigm is well-suited to scenarios with well-defined problem structures and dynamically varying budgets, enabling progressive strategy upgrades as time resources increase.

\item \textbf{Hierarchical (meta-control) combination}: A higher-level strategy serves as a scheduler, dynamically selecting or switching among lower-level sub-strategies based on current task progress or feedback state. For instance, the Soar architecture employs meta-rules to select different sub-strategies according to impasse types, or uses metacognitive heuristics (e.g., ``current strategy has failed too many times'') to trigger strategy switching~\cite{newell1990unified}.

\item \textbf{Parallel (ensemble) combination}: Multiple strategies execute concurrently in different threads or modules, sharing candidate solutions or heuristic information with one another. Under high-concurrency conditions (such as multi-model deployment or parallel branch evaluation in Tree-of-Thought), this paradigm enhances coverage and reduces convergence time. A canonical example is the unified policy-value-search architecture in AlphaZero.
\end{itemize}

This strategic combination mechanism (Figure~\ref{fig:strategy-flowchart}) enables agents to achieve \emph{progressive reasoning} within their time budget: first obtaining a preliminary solution rapidly using low-cost strategies, then gradually introducing deeper, more global, or more creative strategy modules as time progresses. Such \textbf{time-scheduling-driven strategy orchestration} not only improves problem-solving efficiency but also endows the system with greater generality and adaptability.

\subsubsection{Limitations of Mental Models and Individual Preferences}

Johnson-Laird's mental model theory~\cite{johnson1983mental} posits that ``humans do not choose from among all logical possibilities, but rather from a few conceivable models.'' Human reasoning relies on constructing a limited number of mental models. Due to working memory capacity constraints, individuals can typically maintain only three to four models simultaneously, leading to omissions and systematic biases in reasoning~\cite{johnson1999deductive}.

Specific manifestations include:
\begin{itemize}
\item \textbf{Limited model generation}: Inability to enumerate all possible states satisfying given conditions;
\item \textbf{Delayed model updating}: Existing models often cannot be rapidly revised when new information emerges;
\item \textbf{Illusory inference}: Logical judgments are susceptible to the influence of explicit phrasing (e.g., statements containing ``all'' or ``some'' tend to induce logical errors).
\end{itemize}

When confronting a given problem, individuals typically invoke the most familiar strategies that have previously proven effective. This phenomenon can be explained by memory retrieval advantages and principles of cognitive economy: familiar strategies tend to maintain higher activation levels in long-term memory and are readily retrieved when triggered by problem cues~\cite{anderson1990act}. Moreover, under time or resource constraints, individuals preferentially select methods with known lower costs and higher success rates to minimize cognitive load~\cite{rittle1999strategic}. However, over-reliance on familiar strategies may also give rise to the ``Einstellung effect''---persisting with established solution patterns even when the problem context has changed, thereby inhibiting the generation of superior strategies~\cite{luchins1942mechanization}.

\subsubsection{Language and Semantic Reasoning}

Language serves not only as a tool for external human communication but also as a critical mechanism for regulating and organizing thought during cognitive processes. Research in psycholinguistics and cognitive science demonstrates that language is employed not merely to express thought but is also extensively used to construct and manipulate the thinking process itself---a phenomenon particularly evident in the use of inner speech. Inner speech, the process of silently articulating and simulating language internally, is recognized as a cognitive strategy widely deployed by humans during thinking, planning, and reasoning~\cite{vygotsky1986thought}. The existence of inner speech enables individuals to organize and control their own thought processes without relying on external expression.

In \textit{Thinking, Fast and Slow}, Kahneman adopted Stanovich and West's dual-system theory, distinguishing between the fast, intuitive System~1 and the slow, rationally controlled System~2. Although Kahneman did not explicitly address language generation in his exposition of these two systems, it is reasonable to infer that both System~1 and System~2 can drive linguistic output. System~1 typically manifests as automatic, emotional, or reflexive verbal responses (e.g., exclamations, verbal habits), whereas System~2 governs deliberate linguistic expression, such as logical reasoning, planful description, and self-reflection.

More critically, System~2 not only expresses externally through language but also relies substantially on language for internal thinking. As a linguistically mediated form of thought, inner speech plays a central role in higher-order cognitive activities including planning, executive control, and reasoning. Neuroscientific research further indicates that the cognitive resources required by System~2 are primarily localized in the left hemisphere, particularly the prefrontal cortex and language processing centers (e.g., Broca's area). Consequently, language---especially inner speech---can be viewed as a key mechanism for initiating and sustaining System~2, possessing irreplaceable importance in human rational thinking and problem solving.

From a metacognitive perspective, inner speech facilitates self-monitoring of the reasoning process. On one hand, inner speech helps maintain task goals; individuals can sustain focus on target states through self-prompting and internal rehearsal. On the other hand, inner speech enhances monitoring and correction capabilities, enabling individuals to continuously question themselves during thinking, verify the validity of the current path, and make timely adjustments when deviating from objectives.

Empirical research has found that individuals with higher levels of inner speech demonstrate stronger metacognitive regulatory abilities in planning tasks, logical reasoning, and error detection~\cite{alderson2015inner}. This suggests that language not only carries thought but also participates deeply in the thinking process itself.

However, inner speech is likewise constrained by working memory capacity and sustainability. When reasoning paths are complex or information density is excessive, sole reliance on inner speech readily induces cognitive overload. Under such circumstances, partially or fully externalizing the reasoning process (e.g., through writing, dialogue, or structural diagrams) can effectively alleviate cognitive load, enabling thought to unfold in a sustained and controllable manner. This strategy constitutes one of the key mechanisms emphasized in subsequent ``time-enhanced reasoning methods.''

\subsubsection{From Expression to Construction: The Dual Role of Language in Thinking}

In research on the relationship between language generation and thought, two complementary perspectives exist. On one hand, traditional psycholinguistic models (e.g.,~\cite{levelt1989speaking}) view language as an expressive tool for thought, wherein individuals complete conceptual planning before generating language and only subsequently initiate linguistic encoding. This reflects a ``think first, then speak'' cognitive mode. On the other hand, developmental psychology and sociocultural approaches (e.g.,~\cite{vygotsky1986thought}; \cite{dennett1991consciousness}) emphasize the constructive role of language in thought---language not only expresses thinking but actively participates in and shapes reasoning paths during the generation process, manifesting a ``think while speaking'' structural progression mechanism.

In practice, these two mechanisms are often employed alternately across different cognitive tasks: in well-defined tasks and fluent expression, language functions more as the ``output'' of thought; whereas in reasoning, writing, and complex problem solving, language becomes the ``track'' of thought, progressively constructing logic and structure through sequential ordering. Large model training datasets comprise a mixture of language containing reasoning traces alongside non-reasoning linguistic data.

\subsubsection{Problem-Solving Strategies, Language, and Temporal Control}

Although many problem-solving strategies and mental models can operate automatically through procedural memory in everyday human activities~\cite{anderson1982acquisition}, requiring neither explicit control nor linguistic intervention, individuals often need to re-invoke language-based regulatory mechanisms when confronting tasks with complex structures, numerous steps, or ambiguous goals. In such cases, language---particularly inner speech---becomes a key tool for guiding thought, maintaining attention, and organizing strategies~\cite{vygotsky1986thought,alderson2015inner}.

Notably, such linguistic regulatory mechanisms typically exhibit temporal unfolding characteristics: individuals sequentialize multi-step tasks through language, endowing each operation, evaluation, and adjustment with controllable temporal positions and ordering constraints. Through its linear output form, language transforms originally parallel or ambiguous reasoning paths into executable temporal paths, thereby substantially enhancing the stability and schedulability of the problem-solving process. This capacity to ``embed problem solving within the temporal flow'' constitutes a core guarantee for the sustained progression of complex cognitive tasks~\cite{clark1998extended,kahneman2011thinking}.

Inner speech not only helps individuals maintain continuous focus on current subtask goals but also serves as a ``cognitive anchor,'' providing a semantically mediated task maintenance mechanism under limited working memory capacity. Research demonstrates that blocking individuals' inner speech in contexts requiring sustained control and planning significantly impairs their task performance and executive control capabilities~\cite{tullett2010voice}. This indicates that linguistic strategies themselves constitute an effective form of metacognitive regulation, supporting systematic problem space search and strategy evaluation~\cite{flavell1979metacognition}.

Furthermore, externalizing inner speech into written form---such as through writing, drafts, or flowcharts---can effectively alleviate working memory burden, a phenomenon termed ``cognitive offloading''~\cite{morrison2020offloading}. Externalized linguistic structures not only enhance information visibility but also enable strategy monitoring, path reorganization, and escape from local optima~\cite{donald1991origins,norman1993things}.

\begin{figure}[htbp]
	\centering
	\includegraphics[width=0.7\linewidth]{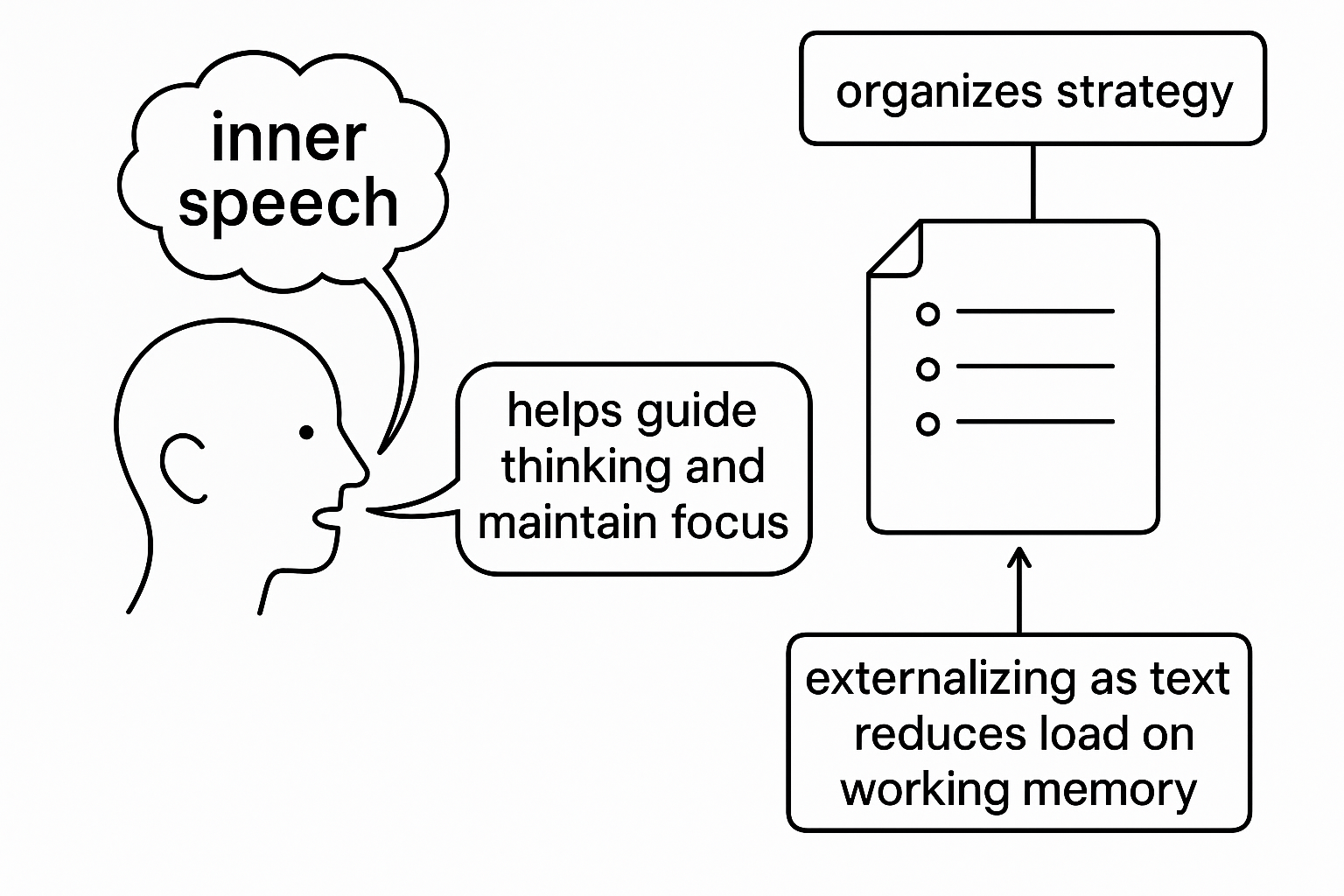}
	\caption{Outputting "inner speech" as text to assist thinking}
	\label{fig:inner-speech}
\end{figure}

For example, when creating a complex presentation or document, individuals may first outline the structure and then fill in details, or alternatively write content while iteratively modifying the structure(Figure~\ref{fig:inner-speech})---the latter psychologically approximates a ``simultaneous offloading and restructuring'' mode of language-driven problem solving embedded within a temporal process. In this context, the outline serves not merely as a planning tool but as an external working memory extension unit~\cite{clark1998extended}, endowing the cognitive process with trackable temporal trajectories and structural explicitness that support subsequent content generation and reasoning elaboration.

Language serves not merely as the output form of problem solving but as its temporalized control structure. Particularly when confronting non-automated, structurally complex problems, language provides a cognitive interface for scheduling, maintenance, updating, and self-supervision. It maintains cognitive stability and mid-course adjustability by organizing the temporal path of reasoning. This mechanism is ubiquitous in human reasoning and has been explicitly simulated in modern reasoning-augmented large models (e.g., DeepSeek-R1), which enhance reasoning consistency and attention maintenance through explicit generation of reasoning traces~\cite{deepseekr1}. The stepwise linguistic output structure in large models is essentially an artificially designed ``temporalized problem-solving process.'' This demonstrates that language is not merely an information carrier but a critical architectural component of cognitive temporality---serving as a key bridge for transferring insights from human cognition to artificial intelligence systems.

\subsection{Artificial Intelligence}

\begin{figure}[htbp]
	\centering
	\includegraphics[width=0.7\linewidth]{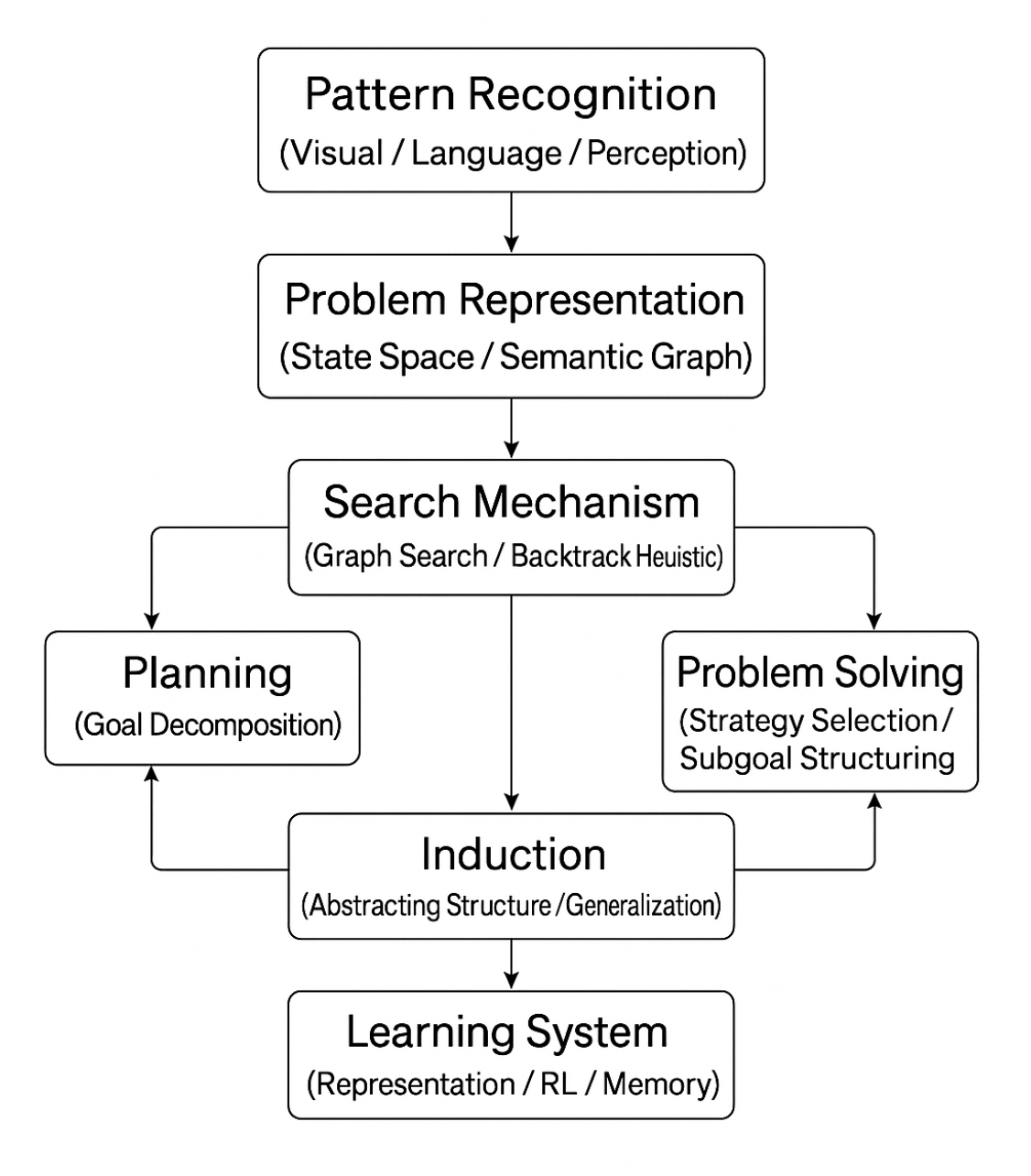}
	\caption{Relationships among the five tasks mentioned in "Steps toward artificial intelligence"}
	\label{fig:minsky-tasks}
\end{figure}

Minsky \cite{minsky1961steps} pointed out in "Steps toward artificial intelligence" that artificial intelligence systems must handle five core tasks: search, pattern recognition, learning, problem solving and planning, and induction. These five capabilities can be expressed independently or interdependently nested within intelligent systems, forming a structural relationship from low-level information processing to high-level cognitive control (Figure~\ref{fig:minsky-tasks}).

These five tasks form an interconnected system. First, \textbf{search} constitutes the most fundamental mechanism in artificial intelligence. Whether finding feasible paths in state space or performing matching and backtracking within knowledge structures, search remains the basic operation connecting inputs to outputs. It serves as the technical foundation for planning and problem-solving behavior.

Second, \textbf{pattern recognition} provides effective problem representations for search by extracting structures, categories, or key features from inputs. For instance, visual systems transform images into states by recognizing shapes and positions; linguistic systems construct semantic structures for reasoning through tokenization, dependency parsing, and related techniques.

\textbf{Learning} represents experiential optimization of the aforementioned processes. Through learning, systems can accumulate pattern recognition preferences, adjust search strategies, and even gradually develop transferable behavioral patterns in environments lacking explicit rules. Methods such as reinforcement learning and representation learning exemplify modern manifestations of this mechanism.

\textbf{Problem solving and planning} constitute the structured integration of search and knowledge with explicit goal orientation. They require not only arranging operation sequences within state space but also dynamically adjusting strategies and subgoal structures. In this process, search depth, strategy selection, and state evaluation are intimately interrelated.

\textbf{Induction} is the process of forming general laws from limited observations. It provides generative hypotheses for learning, initial models for problem solving, and abstract dimensions for planning and evaluation. Induction serves not only as a prerequisite for agents to adapt to their environment but also as the foundation for generalization capabilities.

From a holistic perspective, these five task categories form a cyclical system: recognizing inputs, establishing representations, conducting search, forming solutions, summarizing experience through induction, and continuously optimizing recognition and search strategies through learning.

\subsubsection{Temporal and Structural Credit Assignment in Learning}

In reinforcement learning and neural network training, how to trace outcome responsibility back to prior actions or states constitutes one of the core problems in learning mechanisms. This process is termed ``credit assignment.'' When learning tasks possess temporal extension, systems must not only identify ``what went wrong'' but also determine ``at which step the error occurred'' and ``how much error was incurred.'' Consequently, credit assignment mechanisms must exhibit sensitivity to the temporal dimension.

In reinforcement learning, Temporal Difference (TD) learning represents a canonical algorithm for time-based credit assignment. The TD method updates predictions of future rewards by comparing state value differences between consecutive time steps. This approach combines the long-horizon characteristics of Monte Carlo methods with the recursive properties of dynamic programming, enabling incremental updates without waiting for final outcomes. Its extended form, TD($\lambda$), further introduces the eligibility traces mechanism, allowing credit to propagate backward through time with exponential decay, thereby forming a ``progressive responsibility propagation.''

In neural network training, Backpropagation Through Time (BPTT) serves as the standard optimization method for time-series tasks. It unfolds the standard backpropagation algorithm along the temporal dimension, enabling the model to adjust parameters at multiple past time steps based on future error signals when predicting current output. BPTT can capture long-range dependencies in sequences and constitutes the key technique for training RNN, LSTM, and other sequential models. Since each time step's state may influence the final output, BPTT essentially implements credit distribution backward through time.

Both classes of methods demonstrate the importance of explicitly introducing ``temporal paths'' for responsibility attribution in learning behavior. Compared to structural credit assignment, they emphasize that cognitive systems must learn not only what to do in a given state but also \textit{when} to respond and how to adjust based on \textit{when} effects manifest. This provides the theoretical foundation for subsequent time-enhanced modeling of reasoning processes.

In contrast to temporally unfolded learning mechanisms, another class of credit assignment relies more heavily on structural paths internal to the model. Such mechanisms emphasize layer-by-layer propagation and adjustment of prediction errors or feedback signals along the model's topological structure, achieving fine-grained responsibility attribution within spatial structure.

The backpropagation algorithm represents the canonical example of structural credit assignment. This algorithm employs the chain rule to propagate output errors backward through network layers to each layer's parameters, thereby performing targeted updates to every connection weight in the neural network. In this process, the model does not rely on temporal ordering information between inputs but rather on functional composition relationships between layers in the network structure. This approach achieves precise error distribution across spatial structure, effectively optimizing the model's expressive capacity for single-step tasks.

The advantages of structural credit assignment lie in its high precision and computational stability, making it particularly suitable for tasks such as image recognition and semantic classification, where input structure itself exhibits clear spatial distribution characteristics. In these tasks, the model's objective is to learn mappings from spatial inputs to spatial labels, rather than tracking state evolution across the temporal dimension.

It should be noted that structural credit assignment is not inapplicable to time-series tasks, but without additional mechanisms (such as temporal recurrence or contextual encoding), it struggles to capture cross-timestep causal relationships. Therefore, when handling problems with causal delays and dynamic feedback, structural methods often require combination with temporal mechanisms (such as BPTT or TD learning).

In modern large language models, backpropagation remains the primary training mechanism, but its credit assignment pathways have integrated multi-head attention and feedforward modules from the Transformer architecture, rendering ``structure'' itself highly flexible with certain information flow dynamics.

\subsubsection{Temporal and Structural Reasoning}

In contrast to reasoning mechanisms emphasizing temporal sequence unfolding, another class of artificial intelligence systems focuses on information processing and pattern extraction within spatial or topological structures. Representative examples include static activation networks in the PDP architecture and modern Convolutional Neural Networks (CNNs). The operational mechanisms of such structures do not rely on explicit temporal flow but rather complete reasoning and classification tasks through local adjacency relationships and global hierarchies embodied in the network structure itself.

Within the PDP framework, although time is not directly modeled, information flows between neurons through state activation, constituting a form of ``dynamic propagation within space.'' This mechanism emphasizes connection weights and distributional features between structures rather than sequentiality. Consequently, PDP is better suited for explaining reasoning processes in non-time-sensitive tasks such as visual recognition and associative memory.

CNNs represent a canonical implementation of structure-driven reasoning, widely applied in image processing, object recognition, medical imaging, and related tasks. Through local receptive fields, shared convolutional kernels, and hierarchical abstraction structures, CNNs achieve efficient feature extraction in the spatial dimension. Their reasoning approach more closely resembles biological perceptual systems---performing rapid response and recognition based on the static state of current input without involving sequential memory or long-term state evolution.

Structural reasoning systems are characterized by strong parallelism, fast response speed, and high parameter sharing, making them suitable for processing highly structured or low-time-sensitivity tasks. However, due to their lack of temporal dynamic modeling capability, such systems exhibit limited performance when handling tasks requiring reasoning paths or state changes.

Notably, with the development of neural network architectures, research has attempted to introduce certain temporal mechanisms into structure, such as through dynamic routing, residual connections, and propagation processes in graph neural networks, endowing structural reasoning with certain procedural properties. Nevertheless, structural reasoning fundamentally emphasizes topological relationships between states rather than temporal dependencies between states.

An important class of models in artificial intelligence takes time series as the core modeling object, emphasizing the gradual unfolding of reasoning processes in the temporal dimension. Unlike static rule matching, temporal reasoning emphasizes contextual memory, state updating, and generative prediction within sequences. This mechanism not only reflects the temporal sequentiality of human thought but also constitutes the foundation for language and behavior generation in many cognitive systems.

Although early PDP (Parallel Distributed Processing, 1986) architectures did not explicitly introduce temporal variables, their distributed associative mechanisms can be viewed as an exploration attempting to express time through structure. In PDP networks, information propagation between neurons accumulates ``memory'' through connection strengths and hierarchical structure, with activation patterns evolving as propagation paths change. Temporal factors are ``embedded'' within the network structure in an implicit form---time is encoded as distributed associations between states. This perspective inspired subsequent research paths that model time as a system variable.

The Elman network~\cite{elman1990finding} was the first neural model to explicitly introduce temporal structure. In ``Finding Structure in Time,'' Elman added recurrent connections to feedforward neural networks, making the hidden state at the current time step depend on the state at the previous time step, thereby constituting the most fundamental temporal recurrence mechanism. This model demonstrated the capability to progressively form dynamic representations of syntactic relationships in language processing tasks, proving the critical role of temporal structure in cognitive tasks.

To address the vanishing gradient problem encountered by standard RNNs when processing long-range dependencies, Hochreiter and Schmidhuber~\cite{hochreiter1997long} proposed the LSTM (Long Short-Term Memory) architecture. LSTM extends the model's memory duration for critical information through gating mechanisms, enabling it to handle complex temporal sequences (such as language, music, and action generation). Such models are typically trained using the BPTT (Backpropagation Through Time) algorithm, allowing error signals to propagate backward along the temporal axis and optimize parameters at each time step.

\begin{figure}[htbp]
	\centering
	\includegraphics[width=0.7\linewidth]{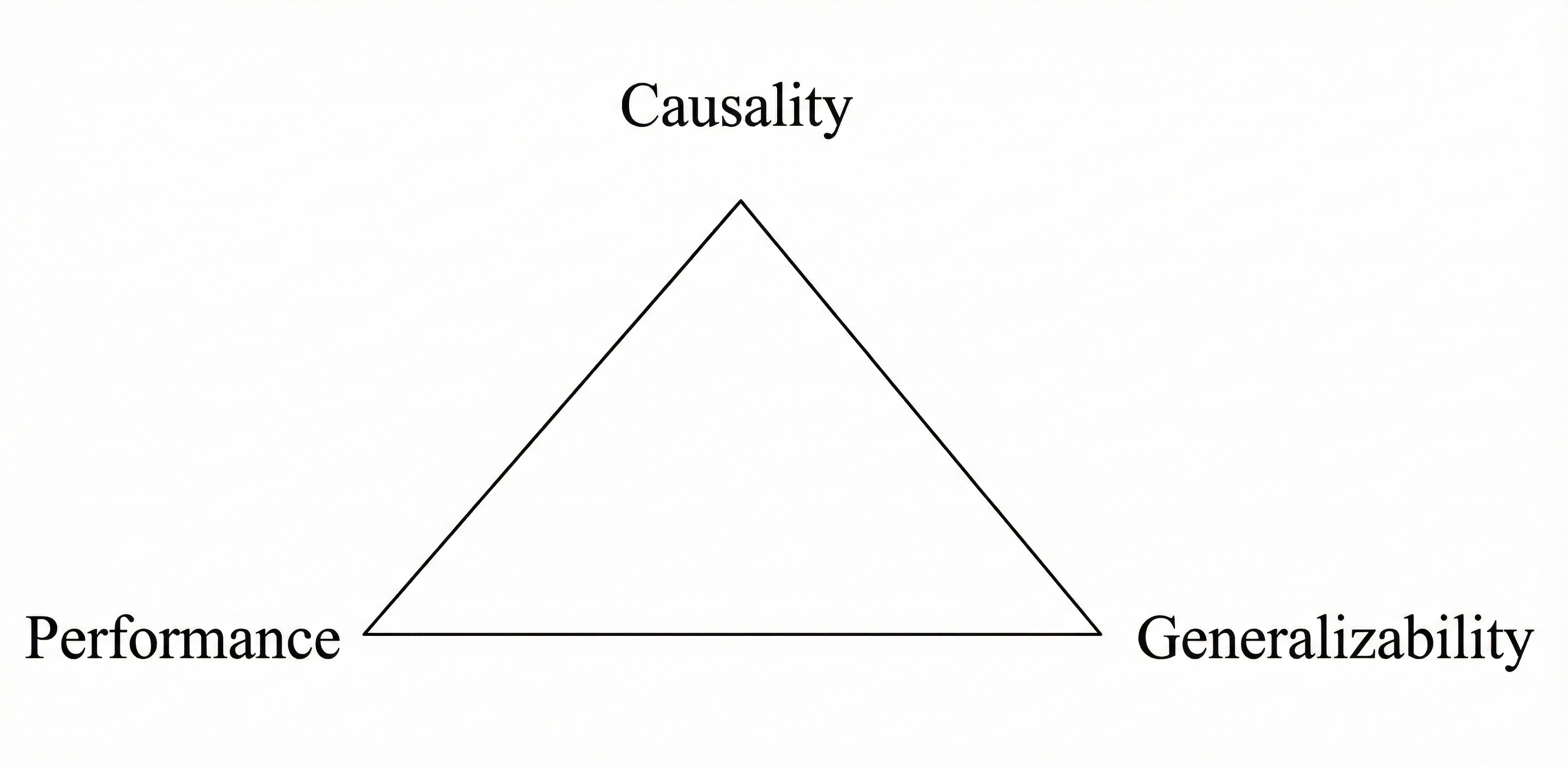}
	\caption{Structural unfolding and temporal unfolding are essentially trade-offs among causality, generalization, and performance. However, for humans, temporal unfolding is constrained by limited cognitive resources. Time is an additional cognitive resource.}
	\label{fig:structure-vs-time}
\end{figure}

Structural unfolding and temporal unfolding represent fundamental trade-offs in cognitive systems (Figure~\ref{fig:structure-vs-time}). With the introduction of the Transformer architecture, internal model structure shifted from sequential unfolding to parallel computation. However, in GPT-class language models, the generation process still retains the core characteristics of temporal reasoning: the model predicts the next token sequentially based on existing context through autoregressive methods. Although recurrent structures are no longer employed, the reasoning process remains a ``time-step-driven'' generation mechanism.

In summary, from PDP's implicit exploration of ``embedding time within structure,'' through RNN and LSTM's temporal state update mechanisms, to GPT's autoregressive generation approach, temporal reasoning mechanisms have consistently constituted the critical path for language understanding and generation tasks. This not only connects behavioral unfolding and semantic evolution in human cognition but has also become an essential foundation for sequence modeling in current artificial intelligence systems.

\section{Temporal Modeling in Reasoning}

At the intersection of cognitive psychology and artificial intelligence, the temporality of reasoning processes has consistently been recognized as an indispensable dimension of intelligent behavior. Reasoning is not an instantaneous static mapping but rather a process of dynamic search within problem space that unfolds over time. Particularly when confronting structurally complex or informationally incomplete problems, reasoning processes require multiple rounds of evaluation, intermediate hypothesis generation, and adjustment---processes that inherently depend on temporal progression.

Temporal reasoning can be modeled at the following levels:

\begin{enumerate}
\item \textbf{Reasoning paths as time series}: Each intermediate conclusion or step serves not only as a unit of knowledge expression but also represents the ``stepping'' of the reasoning process through time. For example, Chain-of-Thought prompting explicitly unfolds reasoning steps, essentially constructing a temporal trajectory that can be carried by language.

\item \textbf{Time budget and reasoning depth}: The reasoning time allocatable to each decision by an agent directly affects its search depth and path complexity. In human cognition, this process typically manifests as the switching mechanism between ``fast thinking'' and ``slow thinking''; in models, it can be controlled by parameters such as token length and step limits.

\item \textbf{Multi-turn interaction and reflection mechanisms}: Allowing agents to perform ``self-review'' after generating preliminary answers or to reconstruct problem representations following user feedback---such processes effectively constitute the unfolding of reasoning over extended temporal ranges. Methods such as Tree-of-Thought and Self-Refine exemplify this mechanism.

\item \textbf{Memory and updating of internal states}: Temporal reasoning requires models to maintain continuous states across different time steps. Early approaches established short-term memory paths through RNN/LSTM, while later GPT series perform weighted integration of all historical tokens through self-attention mechanisms, forming a ``soft temporal sequence modeling'' approach.
\end{enumerate}

Furthermore, temporally unfolded reasoning also relates to the realization of metacognitive capabilities. A system with temporal control capacity can determine ``whether longer reasoning time is currently needed,'' ``whether the current path should be interrupted,'' or ``whether a new thinking branch should be initiated,'' thereby optimizing the scheduling of overall task resources.

Consequently, temporal modeling of reasoning is not merely a technical issue of token generation ordering but rather a manifestation of cognitive strategy scheduling and information control capabilities. In the following sections, we will further explore how to extend, organize, and leverage reasoning time through systematic mechanisms to enhance models' multi-step reasoning capabilities.

\section{Time-Enhanced Large Model Reasoning Methods}

From early prompt strategies like Chain-of-Thought (CoT) and Tree-of-Thought (ToT), to DeepSeek-R1 optimizing reasoning trajectories through reinforcement learning and learning to select appropriate problem-solving strategies for specific problems, the development of reasoning large models shows a transition from heuristic guidance to strategic selection. However, these methods have not substantially changed the ontological structure of model reasoning; the core remains simulating the reasoning process through language output sequences.

By comparison, using more strategy-aware prompts may be more helpful in stimulating the model's problem space search capability. For example, the following prompt:

\textit{"Please act as a problem-solving agent with advanced cognitive capabilities. When facing a complex task, don't be confined to a single strategy. You can flexibly invoke generate-and-test, means-ends analysis, backward reasoning, backtracking, hill climbing, and analogical reasoning strategies, and combine, switch, or nest them according to different stages of the problem. In the solving process, please first clarify the initial state and goal state of the problem, then try to use one strategy to advance. If you encounter difficulties, please actively switch or introduce other strategies to assist. At the same time, please briefly evaluate the effect of each round of strategy, and maintain global control of strategy combination throughout the process. Finally, please summarize which strategy combinations you adopted, which combination was most effective, and whether there is room for further optimization."}

This class of prompts exhibits more pronounced strategic control characteristics than CoT in form; its essence lies in simulating a metacognitive controller at the prompt level, prompting models to dynamically select and combine different strategies during reasoning, thereby enhancing adaptability to open-ended tasks.

However, such methods still face significant limitations: their reasoning processes predominantly rely on single-turn prompts combined with sequential output, lacking explicit evaluation and strategy adjustment mechanisms for intermediate steps. Although methods like ToT formally introduce multi-stage thinking unfolding structures, their overall workflows typically remain enclosed within one-shot generation, failing to form genuinely dynamic interactive processes.

Consequently, if such reasoning mechanisms are further extended to multi-turn dialogue reasoning systems (such as Reflexion, ReAct, etc.), enabling models to perform interactive confirmation, evaluation, and strategy switching with users or the system itself after each thinking step, this would more effectively achieve the enhancement of temporal unfolding and strategic control capabilities in reasoning processes. This not only improves flexibility and adaptability when solving complex tasks but also enables models to progressively acquire characteristics resembling human ``reflective problem solvers.''

From a longer-term perspective, such reasoning mechanisms with interactive self-adjustment capabilities are poised to constitute early prototypes of ``agents with advanced problem-solving capabilities,'' providing a clear pathway embryo for model evolution toward artificial general intelligence.

\section{Discussion and Future Directions}

This paper demonstrates the importance of temporal unfolding from the perspectives of cognitive neuroscience and artificial intelligence development. As computational costs gradually decrease, the problem of humans' inherent limited cognitive resources will not exist in artificial intelligence. Therefore, methods using structure to replace time will still play a significant and decisive role. Nevertheless, agents still need to interact with the environment and humans to obtain more information.

\bibliographystyle{plainnat}
\bibliography{references}

\section*{Appendix A: DeepSeek-R1 Reasoning Process}

\subsection*{Question: "How to solve cognitive fatigue?"}

DeepSeek-R1's reasoning process:

\noindent
"\textcolor{red}{Hmm, the user is asking 'how to solve cognitive fatigue,' which is a common problem faced by modern people.}~\textsc{[MEA]}%
\textcolor{blue}{The user may feel their brain can't work after long hours of work or study, efficiency decreases, or mental exhaustion due to information overload.}~\textsc{[MEA]}%
\textcolor{blue}{The user may be an office worker, student, or professional requiring high brain usage. The direct tone without extra symptom description suggests they may already be aware of being in a state of cognitive fatigue and urgently need practical solutions.}~\textsc{[MEA]}%
\textcolor{green}{The deeper need may not just be temporary relief, but rather establishing a long-term sustainable energy management system.}~\textsc{[MEA]}%
\textcolor{orange}{This question involves multiple levels of neuroscience and psychology. From a physiological perspective, cognitive fatigue is related to brain energy metabolism and neurotransmitter consumption; the psychological level involves attention allocation and emotional regulation.}~\textsc{[ENV]}%
\textcolor{blue}{Need to provide both immediate relief techniques and long-term regulation strategies for the user.}~\textsc{[Generate]}%
\textcolor{gray}{The user may have tried coffee or short breaks with limited effect, so the response should emphasize scientific principles—for instance, explaining why simply ``pushing through'' actually lowers cortisol thresholds, or why multitasking accelerates glucose consumption.}~\textsc{[ENV]}%
\textcolor{blue}{These principles can help users understand the rationale behind the recommendations.}~\textsc{[Test]}%
\textcolor{blue}{Considering implementation difficulty, strategies need to be tiered: use quick solutions like ``3-minute breathing exercises'' in urgent situations, while recommending structured rest methods for daily routines.}~\textsc{[G\&T]}%
\textcolor{blue}{Particularly important to include ``environmental management,'' as many people overlook the implicit impact of lighting and noise on cognitive load.}~\textsc{[ENV]}%
\textcolor{gray}{Finally, it is crucial to alert users to signs of chronic fatigue. Users may normalize fatigue, so it is important to clarify the boundary between pathological and physiological fatigue to avoid delayed medical attention.}~\textsc{[META]}%
\textcolor{gray}{Ending with the rechargeable battery metaphor aims to help users understand that energy fluctuations are a normal mechanism of the human body.}~\textsc{[ANLG]}

\subsection*{Question: "How to travel from Beijing to Shanghai within 5 hours?"}

DeepSeek-R1's reasoning process:

\noindent
"\textcolor{red}{Hmm, the user wants to know how to travel from Beijing to Shanghai within 5 hours.}~\textsc{[MEA]}%
\textcolor{gray}{This is a practical transportation question. The user may need to travel urgently for business or have important matters requiring quick arrival in Shanghai.}~\textsc{[MEA]}%
\textsc{[Generate]}~I see many search results and need to extract the fastest way to reach Shanghai.%
\textcolor{blue}{According to information from web pages 1 and 2, there are two main fast transportation options from Beijing to Shanghai: high-speed rail and airplane. High-speed rail takes approximately 4.5 to 5 hours, while flight time is about 2 hours.}%
\textsc{[Test]}~\textcolor{orange}{Specifically regarding high-speed rail information, web pages 6 and 7 provide detailed train schedules:%
Train G1 departs at 07:00 and arrives at Shanghai Station at 11:29, taking only 4 hours and 29 minutes;%
Train G3 departs at 08:00 and arrives at Shanghai Station at 12:32, taking 4 hours and 38 minutes;%
Train G5 departs at 09:00 and arrives at Shanghai Hongqiao Station at 13:37, taking 4 hours and 37 minutes.%
All these trains complete the journey within 5 hours.}%
\textcolor{orange}{Regarding flights, web pages 8 and 9 show that flight time from Beijing to Shanghai is approximately 2 hours,%
but considering travel to and from airports, security checks, and boarding time, the total duration may be comparable to high-speed rail. Web page 1 mentions that total flight time may be similar to high-speed rail duration.}%
\textsc{[Select]}~\textcolor{green}{Overall, high-speed rail is the most reliable way to reach Shanghai within 5 hours, particularly trains G1, G3, and G5.}%
\textcolor{gray}{Although flights have shorter flight times, when including travel to and from airports and waiting time, the total duration may exceed 5 hours.}%
\textcolor{blue}{In my response, I will emphasize high-speed rail options, particularly those trains that arrive within 5 hours;%
I will create a table displaying recommended train information, including train numbers, departure times, arrival times, total duration, and ticket prices;%
I will also briefly mention flight options but note the risk that total time may exceed 5 hours;%
I will use icons for trains and airplanes to enhance readability and bold key information for emphasis.}~\textsc{[G\&T]}

\subsection*{Color Legend}

\begin{table}[ht]
	\centering
	\renewcommand{\arraystretch}{1.2}
	\begin{tabular}{>{\centering\arraybackslash}m{2.2cm} l}
	  \toprule
	  \textbf{Color} & \textbf{Corresponding Problem Space Element} \\ \midrule
	  \textcolor{red}{\rule{1.6em}{1.6em}}    & Initial State \\
	  \textcolor{blue}{\rule{1.6em}{1.6em}}   & Operators / Action Options \\
	  \textcolor{orange}{\rule{1.6em}{1.6em}} & Intermediate States / Search Path Nodes \\
	  \textcolor{green}{\rule{1.6em}{1.6em}}  & Goal State / Expected Result \\
	  \textcolor{gray}{\rule{1.6em}{1.6em}}   & State Space Structure or Constraints \\
	  \bottomrule
	\end{tabular}
	\caption{Color legend: Corresponding to Problem Space Theory elements}
\end{table}

\begin{table}[ht]
	\centering
	\renewcommand{\arraystretch}{1.2}
	\begin{tabular}{l l}
	  \toprule
	  \textbf{Tag} & \textbf{Corresponding Problem-Solving Strategy} \\ \midrule
	  \textsc{[MEA]}      & Means-Ends Analysis \\
	  \textsc{[G\&T]}     & Generate-and-Test (general) \\
	  \textsc{[Generate]} & Candidate Solution Generation Phase \\
	  \textsc{[Test]}     & Sequential Evaluation Phase \\
	  \textsc{[Select]}   & Optimal Solution Selection Phase \\
	  \textsc{[ANLG]}     & Analogical Reasoning \\
	  \textsc{[META]}     & Metacognitive Strategy Control \\
	  \textsc{[ENV]}      & Problem Space Re-representation \\
	  \bottomrule
	\end{tabular}
	\caption{Text tag explanation: Corresponding to problem-solving strategies}
\end{table}

\end{document}